\documentclass[letterpaper, 10 pt, conference]{ieeeconf}
\IEEEoverridecommandlockouts
\usepackage{cite}
\usepackage{amsmath,amssymb,amsfonts}
\usepackage{algorithmic}
\usepackage{graphicx}
\usepackage{textcomp}
\usepackage{xcolor}
\usepackage{algorithm}
\usepackage{array}
\usepackage[caption=false,font=normalsize,labelfont=sf,textfont=sf]{subfig}
\usepackage{stfloats}
\usepackage{url}
\usepackage{verbatim}
\usepackage{times}
\usepackage{epsfig}
\usepackage{float}
\usepackage{wrapfig}
\usepackage{bm,xspace}
\usepackage{comment}
\usepackage{multirow}
\usepackage{balance}
\usepackage{booktabs}
\usepackage{etoolbox,siunitx}
\usepackage{calc}
\usepackage{pifont,hologo}
\usepackage{nicefrac}
\usepackage{makecell}
\usepackage{svg}
\usepackage{enumerate}

\makeatletter
\let\NAT@parse\undefined
\makeatother
\usepackage[colorlinks=black,citecolor=green]{hyperref}

\usepackage{adjustbox}
\def\BibTeX{{\rm B\kern-.05em{\sc i\kern-.025em b}\kern-.08em
    T\kern-.1667em\lower.7ex\hbox{E}\kern-.125emX}}
    
\def\mathunderline#1#2{\color{#1}\underline{{\color{red}#2}}\color{black}}
\begin{document}

\title{LATITUDE: Robotic Global Localization with Truncated Dynamic Low-pass Filter in City-scale NeRF \\
}

\author{Zhenxin Zhu$^{1,2*}$, Yuantao Chen$^{1,3*}$, Zirui Wu$^{1,4}$, Chao Hou$^{1,5}$, Yongliang Shi$^{1\dag}$,\\ Chuxuan Li$^{1}$, Pengfei Li$^{1}$, Hao Zhao$^{1}$, Guyue Zhou$^{1}$ 
\thanks{*Equal contribution, \dag  Corresponding author}
\thanks{$^{1}$Institute for AI Industry Research (AIR), Tsinghua University, China,
    \{shiyongliang, lichuxuan, lipengfei, zhaohao, zhouguyue\} @air.tsinghua.edu.cn.}%
\thanks{$^{2}$Beihang University, China,
       zhuzhenxin@buaa.edu.cn.}%
\thanks{$^{3}$Xi'an University of architecture and technology, China,
      yuantao@xauat.edu.cn.}%
\thanks{$^{4}$Beijing Institute of Technology, China,
       wuzirui@bit.edu.cn.}%
\thanks{$^{5}$The University of Hong Kong, China,
       houchao@connect.hku.hk.}%
}
\makeatletter
\let\NAT@parse\undefined
\makeatother
\makeatletter
\g@addto@macro\@maketitle
{
  \begin{figure}[H]
  \setlength{\linewidth}{\textwidth}
  \setlength{\hsize}{\textwidth}
  \vspace{-4mm}
  \setcounter{figure}{0}  
  \centering
  \begin{tabular}{@{}c@{\hspace{1mm}}c@{}}
 	\includegraphics[width=0.95\textwidth]{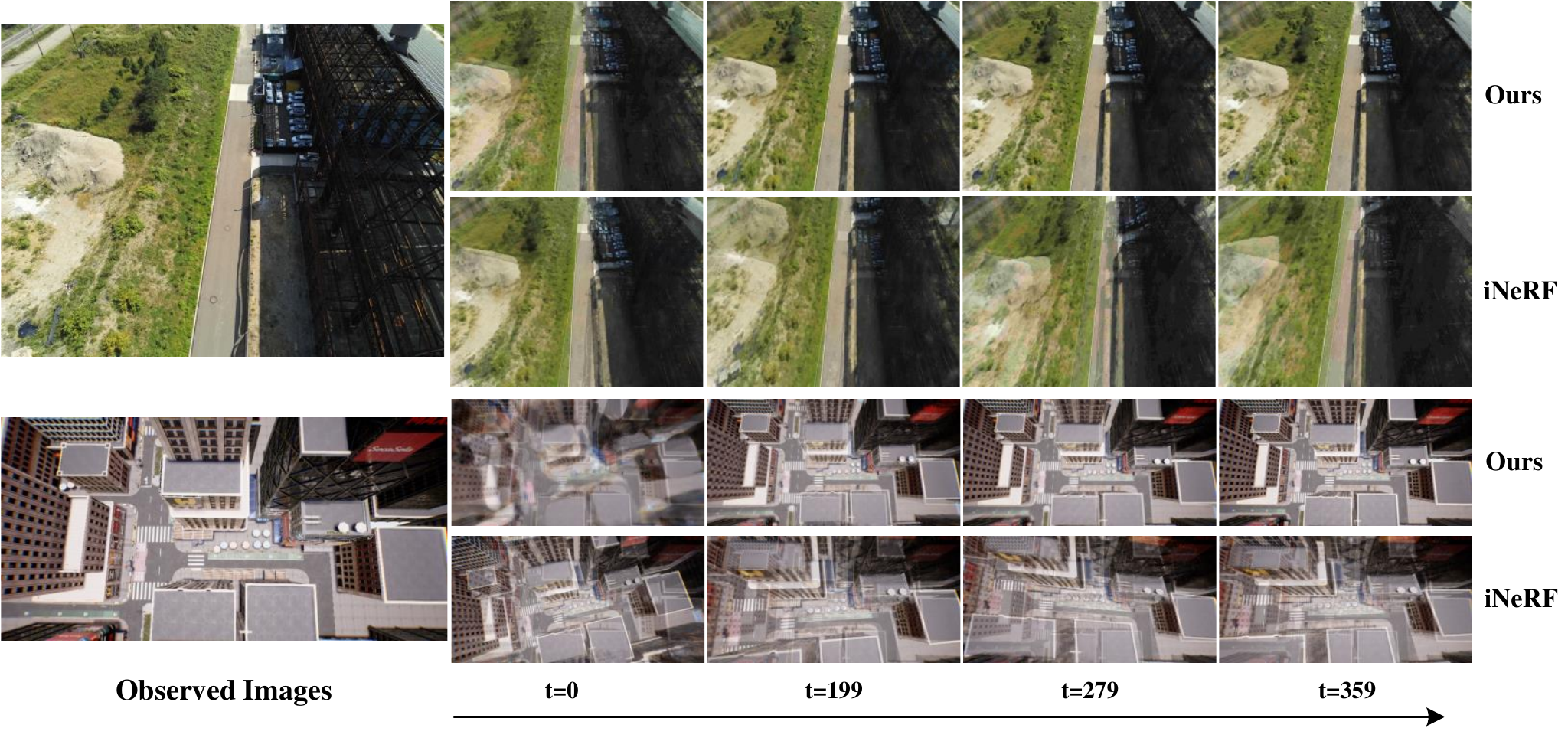}
  \end{tabular}
  \vspace{-2mm}
  \caption{When navigating in city, a new observation is rendered by NeRF for each predicted pose. During iterative optimization, the rendered observations at given positions are updated. With a coarse-to-fine optimization strategy, although a worse initial value is given, our optimization method converges to the exact position and obtains the picture that is almost identical to the actual observation in both real-world (top) and simulation (bottom) scenes.
  }
  \label{fig:teaser}
  \vspace{-4mm}
  \end{figure}
}
\makeatother
\maketitle

\begin{abstract}
Neural Radiance Fields (NeRFs) have made great success in representing complex 3D scenes with high-resolution details and efficient memory. Nevertheless, current NeRF-based pose estimators have no initial pose prediction and are prone to local optima during optimization. In this paper, we present LATITUDE: G\mathunderline{red}{l}obal Loc\mathunderline{red}{a}liza\mathunderline{red}{ti}on with \mathunderline{red}{T}r\mathunderline{red}{u}ncated \mathunderline{red}{D}ynamic Low-pass Filt\mathunderline{red}{e}r, which introduces a two-stage localization mechanism in city-scale NeRF. In place recognition stage, we train a regressor through images generated from trained NeRFs, which provides an initial value for global localization. In pose optimization stage, we minimize the residual between the observed image and rendered image by directly optimizing the pose on the tangent plane. To avoid falling into local optimum, we introduce a Truncated Dynamic Low-pass Filter (TDLF) for coarse-to-fine pose registration. We evaluate our method on both synthetic and real-world data and show its potential applications for high-precision navigation in large-scale city scenes. Codes and dataset will be publicly available at \href{https://github.com/jike5/LATITUDE}{https://github.com/jike5/LATITUDE}.
\end{abstract}


\section{Introduction}
Global camera localization is an essential prerequisite for autonomous navigation tasks. 
Previous APR-based (Absolute Pose Regression) methods \cite{Kendall_2015_ICCV} \cite{DBLP:journals/corr/abs-2110-06558} \cite{https://doi.org/10.48550/arxiv.2204.00559} implicitly match the scene landmarks with image features, which predict rough location from input images but are prone to poor accuracy.
Map-based methods localize the global location of a given observation with explicit maps \cite{caballero2021dll} \cite{chalvatzaras2022survey}, including point cloud, grid-based and mesh-based maps, etc. However, these methods struggle to encode detailed scene appearance using the reasonable disk and memory consumption and thus are limited in generalizing to accurate global localization in city-scale scenes.

On the other hand, the success of NeRF \cite{mildenhall2020nerf} and its follow-up works \cite{Turki_2022_CVPR} represent 3D scenes with neural implicit functions and can render photo-realistic images under arbitrarily given perspectives in the scenes, which offer a possibility to further extend the accuracy of global localization methods by explicitly matching pixel-wised misplacement via stochastic gradient descent (SGD).
Recently, there has been a significant advancement in the field of NeRF-based state estimation \cite{adamkiewicz2022vision} \cite{yen2021inerf} which iteratively optimizes the predicted camera poses and minimizes the photometric errors. Current works, however, are primarily restricted to indoor scenes that are small in scale and in good illumination condition, which are rich in joint visibility and facilitate convergence. However, generalizing  current methods to larger scenes introduces limitations in localization accuracy. This is because, in large-scale scenes, it is hard to find the co-visibility between distant viewpoints, which causes significant difficulties in optimization-based methods.


As mentioned above, optimization-based methods can localize accurate locations but have strict constraints on co-visibility between initial and actual viewpoints, while APR-based methods predict roughly located camera poses, which may potentially provide a good initialization to bootstrap the optimization. To this end, we propose the first method that combines an APR model with a city-scale NeRF for accurate global localization.

%

Our method have two stages: (1) in the first stage, referred to as the place recognition stage, we train an absolute pose regressor that maps the observed images to their rough global locations; (2) in the second stage, i.e., the pose optimization stage, we iteratively optimize the predicted poses based on a pre-trained large-scale NeRF \cite{Turki_2022_CVPR}. During the place recognition stage, we leverage the fact that a large-scale NeRF-based map is available and thus can generate pseudo-pose-image pairs to augment original training data. In the pose optimization stage, we propose to apply a truncated dynamic low-pass filter (TDLF) on the NeRF's positional encoding, which applies a smooth mask on the encoding at different frequency bands (from non-zero to full) over the course of optimization. This coarse-to-fine optimization strategy avoids the local optimum caused by high-frequency information.

To summarize, our contributions are as follows:
\begin{itemize}
\item[$\bullet$]  A two-stage global localization mechanism  in city-scale NeRF is achieved.
\begin{enumerate}[a)]
    \item NeRF-assisted place recognition is achieved, which provides a reliable initial value for pose optimization.
    \item The TDLF is proposed to realize the coarse-to-fine optimization for the initial pose obtained by our NeRF-assisted place recognition.
\end{enumerate}
\item[$\bullet$] We release a dataset for the validation of NeRF-based localization in city-scale simulation.
\end{itemize}

\begin{figure*}[!t]
\centering
\includegraphics[width=0.8\textwidth]{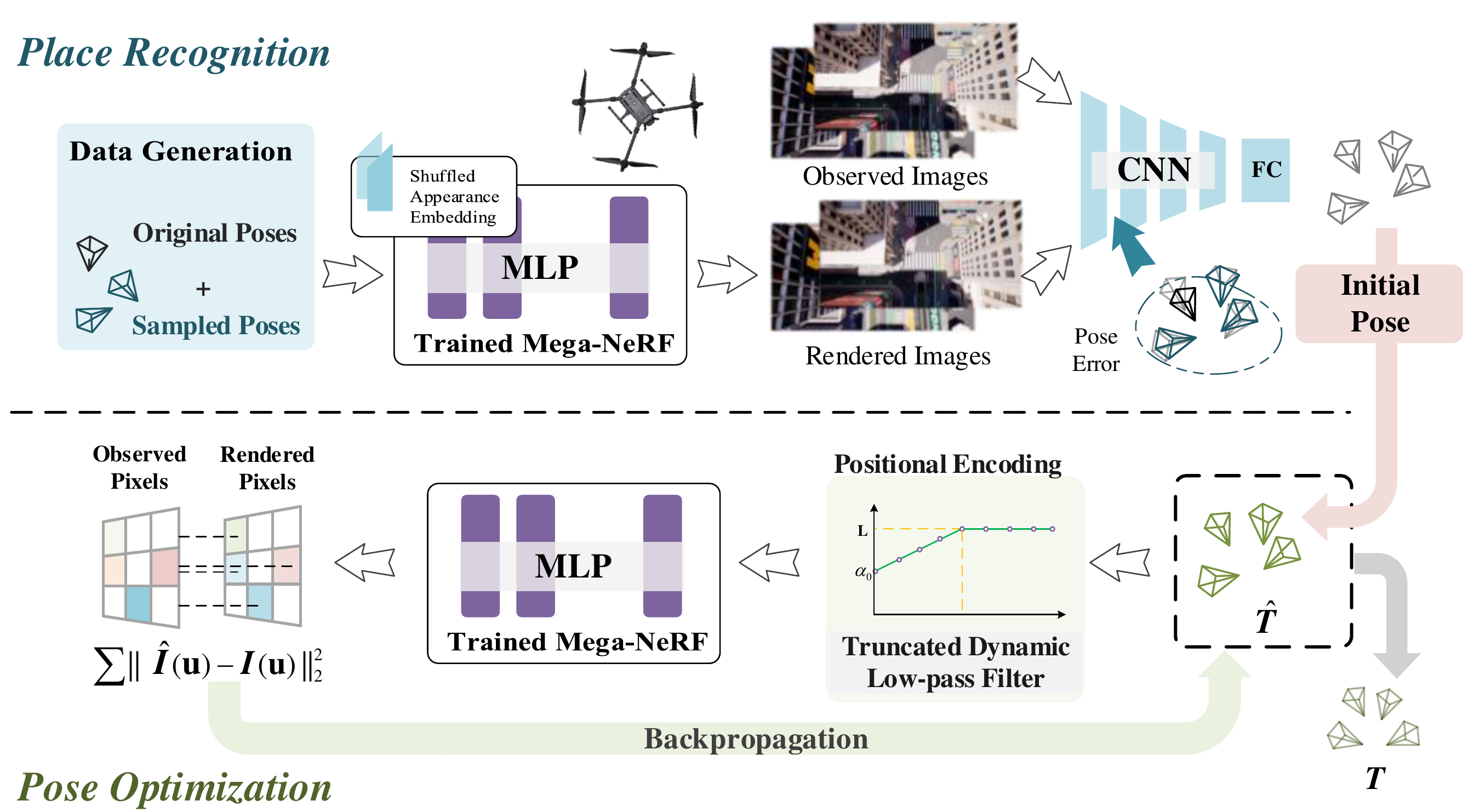}%
\caption{The place recognition module uses camera poses to train the pose regressor. Firstly, additional camera poses are sampled around the original camera poses. Then, the pose vector is passed through Mega-NeRF with shuffled appearance embeddings. Subsequently, the initial poses of the inputted images are predicted by a pose regressor network. During optimization process, on the basis of initial pose, we render the images through Mega-NeRF and backpropagate the photometric error to get a more accurate pose  with the TDLF.}
\label{system}
\vspace{-2mm}
\end{figure*}

\section{Related Work}

\subsection{NeRF-based Representations For Large Scene }
When NeRF is applied to large-scale 3D scene representation, issues arise such as changes in brightness, difficulties in external dynamic object modeling, and unbalanced rendering of the foreground and background in unconstrained scenes.
Martin-Brualla et al. \cite{martinbrualla2020nerfw} used per-frame latent codes to eliminate the difference in brightness and dynamic object appearances.
Zhang et al. \cite{kai2020nerf++} solved the imbalance problem of NeRF in foreground and background rendering by training two MLPs respectively.
Recently, several approaches \cite{Tancik2022BlockNeRFSL}\cite{Turki_2022_CVPR}\cite{xiangli2022bungeenerf} have successfully applied NeRF to the implicit reconstruction of city-scale scenes. Among them, Mega-NeRF \cite{Turki_2022_CVPR} achieves cutting-edge results in novel view synthesis by dividing the map into cells in the spatial domain and fitting a set of NeRFs in parallel to represent the whole scene.
Since NeRF does not constrain volume density distribution during training, NeRF-based surface reconstruction typically encounters problems such as inaccuracy and floating objects. A feasible solution is to introduce additional supervision, e.g. depth maps \cite{kangle2021dsnerf} \cite{roessle2022depthpriorsnerf} and normal priors\cite{wang2022neuris}.


\subsection{Place Recognition}
Place Recongnition (PR) is crucial for the performance of a robot's navigation. A typical pipeline for the PR includes feature extraction, image retrieval and/or 2D-3D feature matching \cite{DBLP:journals/corr/abs-1911-11763} \cite{ sarlin2021back} \cite{7572201}. However, its high cost in terms of computation and memory footprint leaves room for improvement. PoseNet \cite{Kendall_2015_ICCV} proposes the method called absolute pose regression (APR) which first addresses this problem using a deep neural network.
Still, in large-scale scenes, it is difficult to cover the entire scene from one single perspective, so optimizing poses at any position is especially important. LENS \cite{DBLP:journals/corr/abs-2110-06558} has validated that introducing NeRF for data augmentation improves the localization accuracy. However, its training phase is still not memory-efficient. Direct-PoseNet \cite{DBLP:journals/corr/abs-2104-04073} and DFNet \cite{https://doi.org/10.48550/arxiv.2204.00559} deliver the state-of-the-art performances.
\subsection{Pose Estimation with NeRF}
Continuous scene representation makes it possible to recover the camera position of a novel view from a pretrained NeRF. iNeRF \cite{yen2021inerf} presents the first framework to directly optimize 6DoF poses with fixed network parameters by minimizing the photometric residual between rendered pixels and observed pixels. \cite{barf} \cite{nerf--} then success in jointly optimizing camera parameters and scene representations, enabling training NeRF under inaccurate poses. Meanwhile, BARF \cite{barf} introduces a coarse-to-fine pose registration to escape the suboptimal solutions caused by high-frequency positional encoding. Further works \cite{imap} \cite{nice-slam} simultaneously optimize poses and implicit decoder under time-series image sequences. iMAP\cite{imap} shows the first real-time NeRF-based SLAM in the indoor environment. It has a similar pipeline to traditional visual SLAM but substitutes visual features with a single MLP in both tracking and mapping threads. Given RGB-D image sequences, iMAP tracks frames by optimizing poses like iNeRF \cite{yen2021inerf} and selects keyframes for global bundle adjustment for scene mapping. However, these methods are restricted to small indoor scenes.


\section{Formulation}

To represent a 3D scene for novel view synthesis, Neural
Radiance Fields (NeRFs) employ neural networks $\mathcal{F}(\varepsilon):(\mathbf{x},\mathbf{d})\longmapsto(\mathbf{c},\sigma)$ to map position $\mathbf{x}$ and ray direction $\mathbf{d}$ to an emitted color $\mathbf{c}$ and density $\sigma$. We propose a robotic global localization mechanism in a city-scale scene represented by NeRF. Given an observed image, our APR model will predict an initial pose. Then the accurate pose is optimized by the NeRF.
In essence, our localization mechanism is considered as Maximum A Posterior (MAP) Estimation problem. Obtained a prior by place recognition $p( {\hat{T_{k}}|{I_k}})$, we check against the implicit map (Mega-NeRF) in the light of observations to update the prior, and derive our posterior belief about the current location, as is described in eq. \eqref{eq1}. 
\begin{equation}
\label{eq1}
p\left( {\hat{T_k}|\mathcal{F}(\varepsilon),{I_k}} \right) \propto {p\left( {\hat{I_k}|\hat{T_k},\mathcal{F}(\varepsilon)} \right)}   {p\left( {\hat{T_{k}}|{I_k}} \right)} 
\end{equation}
Here, the ${I_{\rm{k}}}$ represents the visual observation at time $k$, and $\hat{I_k}$ is an image rendered by our Mega-NeRF $\mathcal{F}(\varepsilon)$. To achieve an accurate state $T_k$, we optimize the posterior $p( {\hat{T_k}|\mathcal{F}(\varepsilon),{I_k}})$ iteratively by minimizing the photometric loss between the observation $I_k$ and generated image $\hat{I_k}$ by the Mega-NeRF within fixed parameters $\varepsilon$.



\section{Method}

\subsection{System Overview}
As is shown in Fig.\ref{system}, the system includes two modules: place recognition and pose optimization. We first train a NeRF to implicitly model a city-scale scene. The next, combined with actual observed images, a set of images are generated through the Mega-NeRF to train a pose regressor for place recognition. After an initial pose is obtained by the trained regressor, the Mega-NeRF with fixed parameters can optimize it through backpropagation according to loss between rendered and observed images. During optimization, a TDLF is introduced to achieve coarse-to-fine registration.
\subsection{Place Recognition}
Place recognition\cite{Kendall_2015_ICCV} is a prevailing method to address the problem of global localization. In our system, an Absolute Pose Regression method (APR) is adopted to provide reliable initial values at any location in the scene. Our work is divided into two parts: data augmentation and network architecture.

\textbf{Data Augmentation:} Data augmentation is useful to improve performance and outcomes of learning models\cite{shorten2019survey}. In this work, a set of poses sampled near original poses are passed through to the trained Mega-NeRF, and the generated images are considered as data enhancement. Due to the factor that altitude is usually constant in a conventional flight path. To begin, we uniformly sample several positions in a horizontal $H \times W$ rectangle area around each position in training set where $H$ and $W$ is a given parameter.
Then, we need to define the camera orientation attached to these positions for each virtual camera pose.
To avoid degenerate views, we copy the pose orientations of the training set, add random perturbations on each axis drawn evenly in $[-\theta, \theta]$, where $\theta$ is the maximum amplitude of the perturbation. 
Besides, to avoid memory explosion, we generate the poses using the method above and use Mega-NeRF to render images during specific epochs of pose regression training. Additionally, Mega-NeRF's appearance embeddings are selected by randomly interpolating those of the training set, which can be considered as a data augmentation technique to improve the robustness of the 
APR model under different lighting conditions.

\textbf{Network Architecture:} Built on top of VGG16's \cite{vgg16} light network structure, we use 4 full connection layers to learn pose information from image sequences. Given an input image $I_{real}$ and its corresponding ground truth pose $T_{real}(\mathbf{x}_{real}, \mathbf{q}_{real})$, we first use the generated virtual poses $T_{syn}(\mathbf{x}_{syn}, \mathbf{q}_{syn})$ and render a set of virtual images $I_{syn}$. Then the two types of images are inputted into the regressor to get estimated poses $T_{syn}(\mathbf{\hat{x}}_{syn},\mathbf{\hat{q}}_{syn})$ and $T_{real}(\mathbf{\hat{x}}_{real},\mathbf{\hat{q}}_{real})$. 
Finally, our loss functions shown as in eq. \eqref{loss}, and to balance them we introduce two scale factors, where $\gamma$ keeping the expected value of position and orientation error equally, and $\beta$ ensuring the quality of rendered images. In general, the model should trust more on real-world data and learn more from it.
\begin{equation}
\label{loss}
\begin{array}{l}
    \mathcal{L}_{real}=\|\hat{\mathbf{x}}_{real}-\mathbf{x}_{real}\|_{2}+\gamma\left\|\hat{\mathbf{q}}_{real}-\frac{\mathbf{q}_{real}}{\|\mathbf{q}_{real}\|}\right\|_{2} \\
    \\
    \mathcal{L}_{syn}=\|\hat{\mathbf{x}}_{syn}-\mathbf{x}_{syn}\|_{2}+\gamma\left\|\hat{\mathbf{q}}_{syn}-\frac{\mathbf{q}_{syn}}{\|\mathbf{q}_{syn}\|}\right\|_{2}\\
\end{array}
\end{equation}
\begin{equation}
\label{factor}
    \mathcal{L}=\mathcal{L}_{real} + \beta\mathcal{L}_{syn}
\end{equation}

\subsection{Pose Optimization}
The initial pose obtained from the place recognition is limited in terms of precision. In order to further improve the accuracy, we optimize pose on tangent plane to ensure a smoother convergence on one hand, and on the other hand by implementing the TDLF can avoid from falling into the local optimum during optimization course.

\textbf{Optimization on Tangent Plane:} 
Generally, gradient-based
optimization on SE(3) is utilized to solve for the pose estimation\cite{yen2021inerf}.
However, as mentioned in \cite{adamkiewicz2022vision}, optimization on the tangent plane can performs smoother and quicker convergence. We use the same approach as in \cite{adamkiewicz2022vision} and formulate the problem as eq. \eqref{eq3}, where ${I}$ is an observed image and $\hat{T}_0$ is the associated initial pose.
\begin{equation}\hat{\xi } = \mathop{\arg\min}\limits_{\xi \in \mathfrak{se}\left(3 \right) }{\mathcal{L}(\xi | \hat{T}_0, {I}, \theta) }\label{eq3}\end{equation}
We use eq. \eqref{eq4} to obtain the optimal pose estimation, where $\hat{\xi}$ is the optimization variables.
\begin{equation}\hat{T}=\exp(\hat{\xi}) \hat{T}_0\label{eq4}\end{equation}

To further demonstrate our loss function $\mathcal{L}$, we introduce the 3D world coordinates, where $T$ denotes the transformation from camera coordinates to the world coordinates. Then given some sampled set ($\mathcal{R}$) of pixel coordinates: $\mathbf{u} \in \mathcal{R}$, the loss function can be expanded as eq. \eqref{eq5}, where $K$ is camera intrinsic parameters and $\mathcal{I}:\mathbb{R}^{4N} \rightarrow \mathbb{R}^3$ is compositing function. Our goal is to optimize the pose by back propagation of the loss in eq. \eqref{eq5}. 

\begin{equation}
\mathcal{L} = \sum _{\mathbf{u} \in \mathcal{R}} {\left \| \mathcal{I}\left(\mathcal{F}(TK\mathbf{u};\theta) \right ) - I(\mathbf{u}) \right \|^2_2 }
\label{eq5}\end{equation}

\begin{figure}[!t]
\centering
\includegraphics[width=0.5\textwidth]{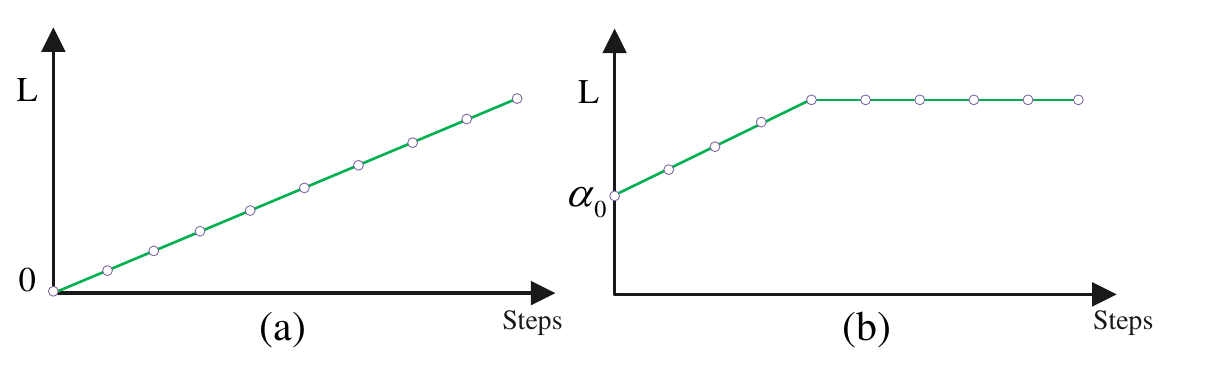}%
\caption{(a) Dynamic Low-pass Filter. (b) Truncated Dynamic Low-pass Filter.  Method (a) starts from zero which will lead to unreasonable inference results, on the contrary, method (b) ensures the validity of the inference information in the initial stage.}
\label{low-pass-filter}
\end{figure}

\textbf{Truncated Dynamic Low-pass Filter:} 
Typically, images are complex signals, which means it's easy to fall into local minima when using gradient descent with photometric loss. In the field of 2D image alignment, a very simple but effective operation is blurring the image first to make the signal smoother and ensure that local minimum can be stepped out. Similarly, \cite{barf} proposed the dynamic low-pass filtering as  as shown in Fig. \ref{low-pass-filter}(a), where the frequency of positional encoding increases from zero to full linearly during iterations. It demonstrated that dynamic low-pass filter can suppress high-frequency signals to achieve the same effect as blurring images.

\begin{figure}[!t]
\centering
\includegraphics[width=0.48\textwidth]{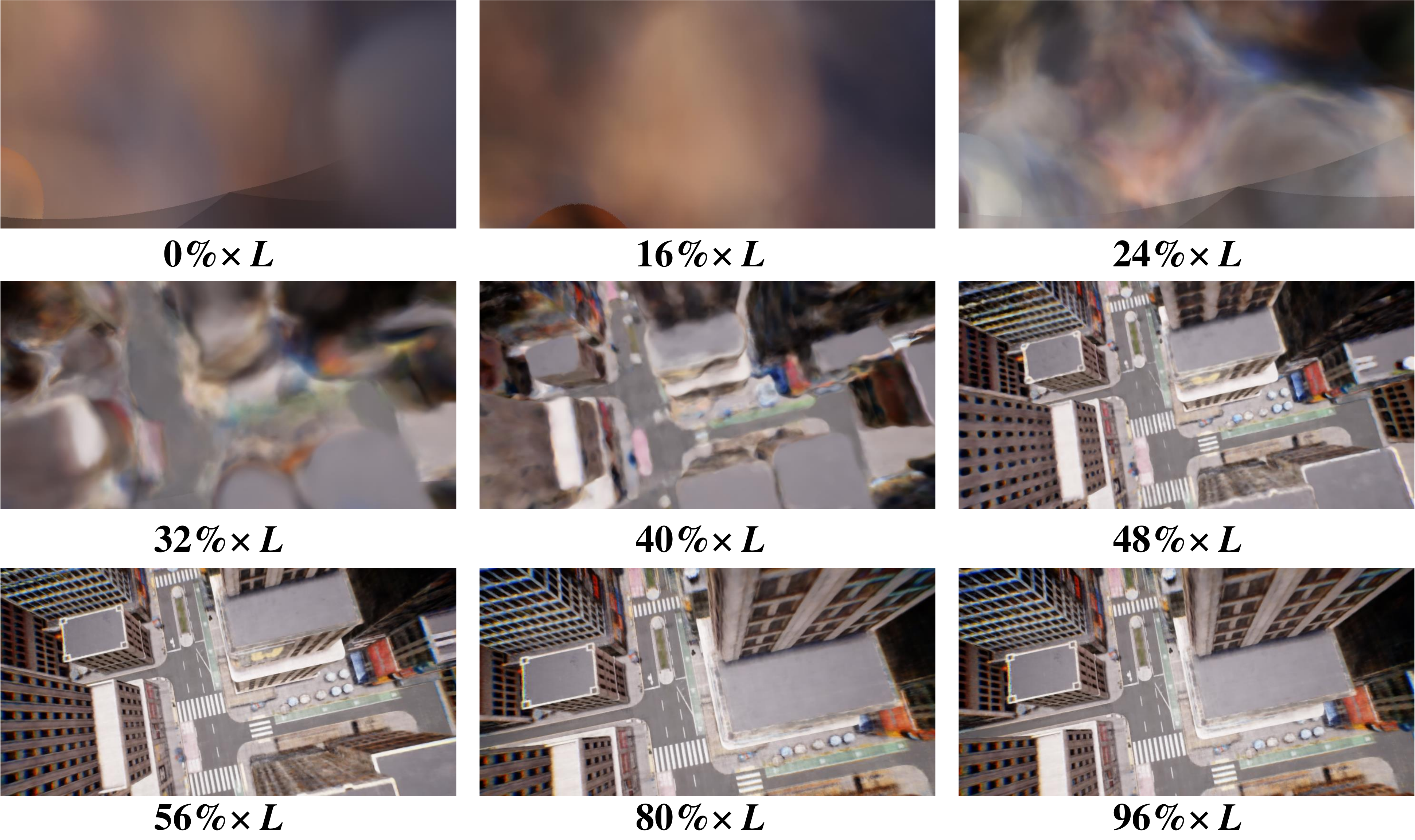}%
\caption{Given a trained NeRF network, the above results are obtained by changing $\alpha$ from $0$ to $L$ dynamically during the inference. It shows that the images rendered before $\alpha=24\% \times L$ contains invalid or even wrong color information, which will lead to pose optimization easily scattered.}
\label{low-pass-filter-imgs}
\vspace{-2mm}
\end{figure}

However, when a trained network is given, it's unnecessary to set the frequency threshold $\alpha$ start from zero, for it will lead to the network infer an invalid or even wrong RGB results at the first place, as shown by the first line in Fig.\ref{low-pass-filter-imgs}. Through experiments we found that the pose estimation diverge easily, due to the inference results contain a lot of invalid appearance. To solve the problem, our TDLF set $\alpha$ starts from $\alpha_0$ to $L$ as shown in Fig. \ref{low-pass-filter}(b), where $\alpha_0$ is a non-zero number, and $L$ is the full positional encoding frequency. By using TDLF, we can obtain more accurate color information at the early stage of optimization to ensure the proper convergence of the pose, while suppressing high-frequency signals as much as possible to avoid falling into local optima.
Formally, the $k$-th positional encoding function we use is eq.\eqref{eq9}, where the weight $\omega_k$ is calculated as in eq. \eqref{eq10} and $\alpha$ is the ratio of the current iteration to the total number of iterations.

\begin{equation}
\gamma_k(\mathbf{x}, \alpha)=\omega_k(\alpha)\cdot [\cos (2^k \pi \mathbf{x}), \sin(2^k \pi \mathbf{x})]
\label{eq9}\end{equation}

\begin{equation}
\omega_k(\alpha) = \begin{cases}
1 &  k \le (\alpha  + \frac{\alpha_0}{L})L \\
0 &  k >   (\alpha  + \frac{\alpha_0}{L})L
\end{cases}\label{eq10}
\end{equation}

\section{Experiments and Results}
\subsection{Datasets}
We evaluate our method against both our own Urban Minimum Altitude Dataset (UMAD) and Mill 19 \cite{Turki_2022_CVPR} dataset. The UMAD is a virtual-scene dataset made by the simulator AirSim \cite{shah2018airsim}, which is built on top of the Unreal Engine. In order to ensure the simulation data is as close to real as possible, on the one hand, we use a realistic city scene model which comes from Kirill Sibiriakov \cite{ArtStation}, on the other hand, we collect the drone trajectory data and image data separately to ensure the frequency and quality derived from methods in the paper \cite{tartanair2020iros}. In comparison to the other urban datasets \cite{brunel:hal-03380109}\cite{UrbanScene3D} which are created using the oblique aerial photography technique, our dataset has higher fidelity when it comes to the texture details.

\begin{figure}[htbp]
\centering
\vspace{-5mm}
\includegraphics[width=0.46\textwidth]{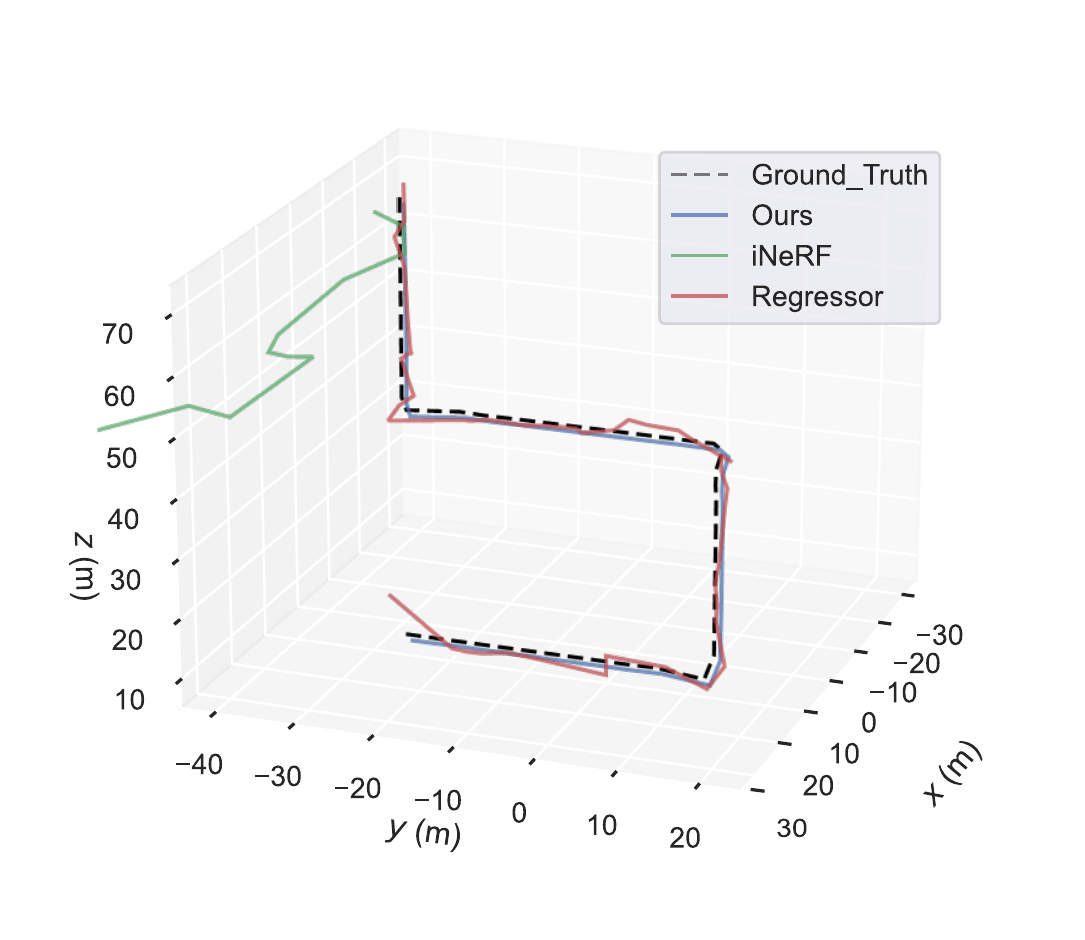}%
\vspace{-4mm}
\caption{Trajectory comparison on partial sequence of UMAD.}
\label{fig:trajcom}
\end{figure}

\subsection{Implementation Details} 
We follow the architectural settings from Mega-NeRF \cite{Turki_2022_CVPR}, still, with several modifications for scene representation.  The synthetic scene is divided into four parts. We train each part of the scene for 300,000 iterations. The submodule MLP consists of 8 layers of 256 hidden units and a final fully connected ReLU layer of 128 channels. We resize the images to 960 × 480 pixels and randomly sample 1024 pixel rays during the training steps. We use an Adam optimizer with an initial learning rate of $5 \times {10}^{-4}$ decaying exponentially to $5 \times {10}^{-5}$. We discard the practice of adding Gaussian noise to the sigma output proposed by the original NeRF \cite{mildenhall2020nerf} because it harms the final results.

To train the place recognition model, we resize the rendered images to 480 × 240 pixels, then we use the CNN model to extract the features. An Adam optimizer is used, whose initial learning rate is $1 \times {10}^{-4}$ decaying exponentially to $2 \times {10}^{-5}$. We set $\gamma=250$ and $\beta=0.8$ to balance the learning of location and orientation, real images and synthetic images.   

In order to implement TDLF in the positional encoding layer, we add the parameter $\omega$ in eq. \eqref{eq10} to the network module of the layer. For less frequent changes of this parameter, we set its value every 50 steps during the pose optimization stage. The initial threshold $\alpha_0$ is set to $40\% \times L$ in the experiments. Experiments are conducted on an NVIDIA RTX3090 GPU with 24GB of memory.

\subsection{Global Localization}
In place recognition stage, the image sequence from the UMAD is given into the trained regressor, outputting the initial pose values. In optimization stage, the initial poses are optimized using our method and iNeRF (in fairness, we replaced its MLP with mega-nerf, and the same was done in ablation study) respectively. As can be seen in Fig. \ref{fig:trajcom}, the prediction of regressor is within error tolerance of our pose optimization method,  and the results are almost consistent with the ground truth, while iNeRF fails to converge. According to the quantitative evaluation results of UMAD scene in Table \ref{tab:traj}, the pose error obtained from the regressor is within the range of $0 \sim 8$ $m$ , and our optimization method can further converges to the exact position. The average error is only less than 0.05 $m$, and the minimum error is even less than 0.01 $m$. The average error of iNeRF is up to 12.36 $m$, and the minimum error is also more than 5 $m$. Our method, compared to iNeRF,  also performs better in the Mill 19.


\begin{table}[t]
\renewcommand\arraystretch{1.3}
\centering
\caption{Quantitative results on global translation error (m).}
\label{tab:traj}
\begin{tabular}{l|llllll}
\hline
Scenes                & Methods   & Max   & Mean  & Min   & Rmse  & Std   \\ \hline
\multirow{3}{*}{UMAD} & Regressor & 7.03  & 1.69  & 0.34  & 2.07  & 1.20  \\
                      & iNeRF     & 56.18 & 12.36 & 4.70  & 18.28 & 13.48 \\
                      & Ours      & \textbf{0.25}  & \textbf{0.05}  & \textbf{0.01}  & \textbf{0.06}  & \textbf{0.04}  \\ \hline
\multirow{3}{*}{\begin{tabular}[c]{@{}l@{}}Mill 19\end{tabular}} & Regressor & 8.77 & 2.22 & 0.39 & 3.13 & 2.20 \\
                      & iNeRF     & 33.81 & 17.52 & 19.42 & 20.43 & 10.51 \\
                      & Ours      & \textbf{0.15}  & \textbf{0.11}  & \textbf{0.06}  & \textbf{0.11}  & \textbf{0.02}  \\ \hline
\end{tabular}
\end{table}

\subsection{Ablation Study Of Optimization}
Our state estimation has improved on the basis of iNeRF. On the one hand, we perform optimization on the tangent plane which attributes to the noisy photometric loss landscape over the SE(3) manifold; on the other hand, we introduce a TDLF to apply a smooth mask on the encoding at different bands (from low to high) over the course of optimization. Therefore, our ablation study will conduct a quantitative analysis of the error anti-interference ability and convergence accuracy of pose estimation according to the improvements. In the absence of TDLF and manifold-based optimization, our method is equivalent to iNeRF, just replacing the original NeRF with Mega-NeRF, which is more suitable for large scenes.

\begin{figure*}[!t]
\centering
\includegraphics[width=0.9\textwidth]{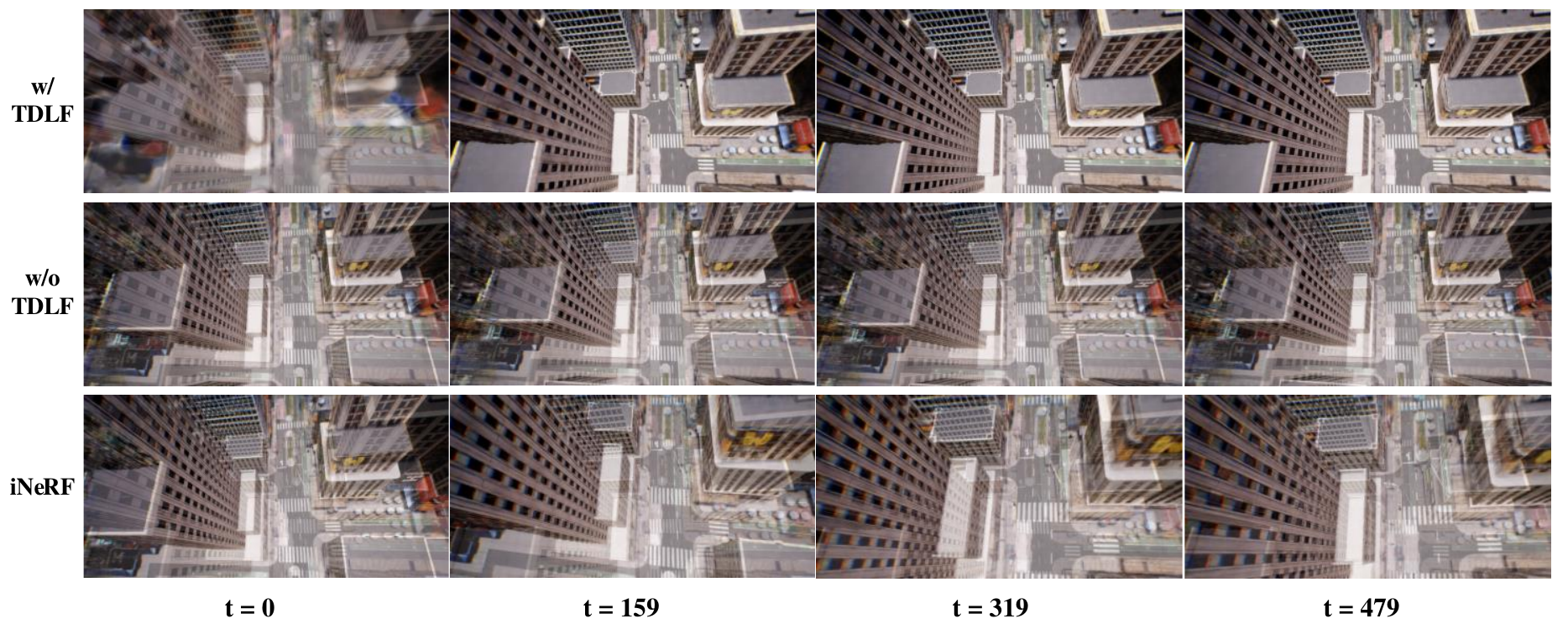}%
\vspace{-1mm}
\caption{We visualize the average results of rendered images in repeated experiments based on the estimated pose at iteration t and test image to compare our model with iNeRF. The utilization of the dynamic low-pass filter enables escaping from local minima, which is shown by the first line. The comparison of the third line and the second line implies that recursively optimizing on SE(3) instead of tangent plane leads to incorrect estimation results.}
\label{fig:ablation}
\vspace{-2mm}
\end{figure*}

We carried out ablation experiments on the UMAD dataset, taking four different positions in the same configuration and recording their average results in Table \ref{tab:transerror} and Table \ref{tab:roterror}. The tables show that, our method has the largest tolerance for both translation and rotation perturbations, and the highest convergence accuracy. Meanwhile, it also proves the effectiveness of the TDLF in the NeRF-based state estimation method. Results in Fig.\ref{fig:ablation} show that, in the initial phase of pose estimation, low-frequency contour information is used for registration, and high-frequency information is gradually used, which is similar to the idea of branch-and-bound. It guarantees a good way to escape from local minima generated by the high-frequency non-convex position encoding, while without TDLF, it is easy to fall into the local optimum. Besides, compared with the direct optimization on SE(3) in iNeRF, the optimization on the tangent plane also shows its superiority. Additionally, when the translation error reaches 16m and the rotation error reaches 16°,
the localization performance of our method also degrades due to the small overlap between the observed image and the rendered image at larger offsets leading to a local optimum.

\begin{table}[!t]
\renewcommand\arraystretch{1.2}
\centering
\caption{Ablation study with different initial translation error.}
\label{tab:transerror}
\begin{tabular}{ccccc}
\hline
\begin{tabular}[c]{@{}c@{}}Initial\\ Error(m)\end{tabular} &
  \begin{tabular}[c]{@{}c@{}}Manifold\\ Optimization\end{tabular} &
  \begin{tabular}[c]{@{}c@{}}TDLF\end{tabular} &
  \begin{tabular}[c]{@{}c@{}}Translation\\ Error(m)\end{tabular} &
  \begin{tabular}[c]{@{}c@{}}Rotation\\ Error(°)\end{tabular} \\ \hline
\multirow{3}{*}{4}  & $\bm{\times}$  & $\bm{\times}$  & 0.19          & 0.69          \\
                    & \checkmark & $\bm{\times}$  & 0.83          & 0.21          \\
                    & \checkmark & \checkmark & \textbf{0.02} & \textbf{0.10} \\ \hline
\multirow{3}{*}{8}  & $\bm{\times}$  & $\bm{\times}$  & 22.78         & 8.81          \\
                    & \checkmark & $\bm{\times}$  & 1.97          & 0.17          \\
                    & \checkmark & \checkmark & \textbf{0.02} & \textbf{0.09} \\ \hline
\multirow{3}{*}{12} & $\bm{\times}$  & $\bm{\times}$  & 7.14          & 4.30          \\
                    & \checkmark & $\bm{\times}$  & 2.10          & 4.47          \\
                    & \checkmark & \checkmark & \textbf{0.03} & \textbf{0.10} \\ \hline
\multirow{3}{*}{16} & $\bm{\times}$  & $\bm{\times}$  & 7.41          & 5.86          \\
                    & \checkmark & $\bm{\times}$  & 4.19          & 2.10          \\
                    & \checkmark & \checkmark & \textbf{3.84} & \textbf{0.97} \\ \hline
\end{tabular}

\end{table}

\begin{table}[!t]
\renewcommand\arraystretch{1.2}
\centering
\caption{Ablation study with different initial rotation error.}
\label{tab:roterror}
\begin{tabular}{ccccc}
\hline
\begin{tabular}[c]{@{}c@{}}Initial\\ Error(°)\end{tabular} &
  \begin{tabular}[c]{@{}c@{}}Manifold\\ Optimization\end{tabular} &
  \begin{tabular}[c]{@{}c@{}}TDLF\end{tabular} &
  \begin{tabular}[c]{@{}c@{}}Translation\\ Error(m)\end{tabular} &
  \begin{tabular}[c]{@{}c@{}}Rotation\\ Error(°)\end{tabular} \\ \hline
\multirow{3}{*}{4}  & $\bm{\times}$             & $\bm{\times}$             & 8.72          & 13.5          \\
                    & \checkmark & $\bm{\times}$             & 0.86          & 1.03          \\
                    & \checkmark & \checkmark & \textbf{0.02} & \textbf{0.09} \\ \hline
\multirow{3}{*}{8}  & $\bm{\times}$             & $\bm{\times}$             & 20.95         & 17.00         \\
                    & \checkmark & $\bm{\times}$             & 3.18          & 5.45          \\
                    & \checkmark & \checkmark & \textbf{0.02} & \textbf{0.10} \\ \hline
\multirow{3}{*}{12} & $\bm{\times}$             & $\bm{\times}$             & 7.81          & 28.92         \\
                    & \checkmark & $\bm{\times}$             & 6.12          & 11.03         \\
                    & \checkmark & \checkmark & \textbf{0.70} & \textbf{3.39} \\ \hline
\multirow{3}{*}{16} & $\bm{\times}$             & $\bm{\times}$             & 9.99          & 19.74         \\
                    & \checkmark & $\bm{\times}$             & 6.12          & 17.67         \\
                    & \checkmark & \checkmark & \textbf{3.24} & \textbf{4.80} \\ \hline
\end{tabular}
\end{table}

\subsection{Effect of Frequency Threshold}
With different frequency threshold $\alpha_0$, we have evaluated our method on both UMAD and Mill 19 dataset with a translation error of 8m and a rotation error of 8°, and Table \ref{tab:wtdlf} shows that our method performs best around a threshold $\alpha_0$ of $40\% \times L$. Here, the TDLF acts as if it were a BARF when $\alpha_0$ is $0\% \times L$. The results indicate that TDLF optimizes for more accurate alignment compared to the baselines. This highlights the effectiveness of TDLF utilizing a coarse-to-fine strategy for localization.

\begin{table}[t]
\renewcommand\arraystretch{1.3}
\centering
\caption{Quantitative results in different degrees of truncation on the whole frequency.}
\label{tab:wtdlf}
\resizebox{0.98\linewidth}{!}{
\begin{tabular}{c|cllllll}
\hline
Scenes                & Error & 0\%   & 10\%   & 30\% & 40\%  & 50\%   & 70\%   \\ \hline
\multirow{2}{*}{UMAD} 
                      &Trans(m)  & 200.07 & 81.18 & 0.04 & \textbf{0.02} & 0.03 & 0.06  \\
                      &Rot(°)    & 34.10 & 50.24 & 28.49 &   \textbf{0.10} & 0.18 & 0.20 \\ \hline
\multirow{2}{*}{\begin{tabular}[c]{@{}l@{}}Mill 19\\ Building\end{tabular}} 
                      &Trans(m)  & 0.69 & 0.18 & 0.09 &     \textbf{0.06} & 0.07 & 0.10  \\
                      &Rot(°)    & 7.92 & 4.83 & 1.97 &     0.20 & \textbf{0.14} & 3.24   \\ \hline
\end{tabular}
}
\end{table}



\section{Conclusion}
In this work, we propose a two-stage global localization mechanism under city-scale NeRF. A pose regressor is trained to provide an initial pose for a robot at arbitrary position. Besides, we introduce a TDLF for optimization to achieve coarse-to-fine pose registration. By conducting extensive experiments on both simulation and real-world dataset, our method, compared to recent works, is superior in both accuracy and tolerance of errors. 


\section*{ACKNOWLEDGEMENTS}
This work was sponsored by Tsinghua-Toyota Joint Research Fund (20223930097).

\bibliographystyle{plain}  
\bibliography{root}

\vspace{12pt}

\end{document}